\documentclass[10pt,twocolumn,letterpaper]{article}

\usepackage{wacv}
\usepackage{times}
\usepackage{epsfig}
\usepackage{graphicx}
\usepackage{amsmath}
\usepackage{amssymb}
\usepackage{enumerate}

\newcommand{\argmin}{\operatornamewithlimits{argmin}}

\wacvfinalcopy 

\def\wacvPaperID{****} 

\ifwacvfinal\fi
\begin{document}

\title{Generalized Adaptive Dictionary Learning via Domain Shift Minimization}

\author{Varun Panaganti\\
Indian Institute of Technology Madras\\
{\tt\small varunpanaganti@gmail.com}
}

\maketitle
\ifwacvfinal\thispagestyle{empty}\fi

\begin{abstract}
Visual data driven dictionaries have been successfully employed for various object recognition and classification tasks. However, the task becomes more challenging if the training and test data are from contrasting domains. In this paper, we propose a novel and generalized approach towards learning an adaptive and common dictionary for multiple domains. Precisely, we project the data from different domains onto a low dimensional space while preserving the intrinsic structure of data from each domain. We also minimize the domain-shift among the data from each pair of domains. Simultaneously, we learn a common adaptive dictionary. Our algorithm can also be modified to learn class-specific dictionaries which can be used for classification. We additionally propose a discriminative manifold regularization which imposes the intrinsic structure of class specific features onto the sparse coefficients. Experiments on image classification show that our approach fares better compared to the existing state-of-the-art methods.
\end{abstract}

\section{Introduction}
The study of sparse representation of signals has received an enormous interest in the recent years. The idea behind sparse representation is to approximate a signal by representing it with a combination of very few elements from an over-complete set of bases called dictionary, i.e. any natural signal can be reconstructed by a sparse combination of elements of an over-complete dictionary. Much of the earlier work on sparse representation was devoted to building a dictionary using off-the-shelf or parametric bases. The notion of building a dictionary from data instead of a predefined set of bases was studied by Olshausen and Field \cite{olshausen1997sparse} in their seminal work. Data driven dictionaries have since yielded encouraging results among tasks like restoration \cite{elad2006image}, super-resolution \cite{yang2012coupled,wang2012semi} and classification \cite{wright2009robust}.

The effectiveness of these dictionaries in such diverse range of applications can be attributed to their superior ability in adapting to a particular set of data. However we might encounter situations in which the target data has a distribution different from the data used in training the dictionary. Such situations occur frequently in many computer vision problems e.g., changes in resolution, illumination and pose of images. Such changes often lead to degradation in classification performance \cite{daume2009frustratingly}. Learning dictionaries which are adaptive to these changes is a challenging task, which has been garnering increased interest of late. Earlier works were focussed on learning a dictionary for each domain. Jia $et$ $al.$ \cite{jia2010factorized} considered such a case. But the dimension of the features is often high, hence learning a dictionary for each domain is cumbersome and computationally expensive, making it infeasible for many practical applications.   


The idea of adapting classifiers to new domains has attracted a tremendous amount of interest recently, and a number \cite{saenko2010adapting,kulis2011you,gopalan2011domain,jhuo2012robust} of methods have been proposed. Jhuo $et$ $al.$ \cite{jhuo2012robust} proposed learning a transformation of source data onto the target space, such that the joint representation is low-rank. However, they do not effectively utilize the labeled data to learn the projections.  Han $et.$ $al$ \cite{han2012sparse} learned a shared embedding for different domains, with a sparsity constraint on the representation. Albeit, they treat the step of embedding the data onto a common domain separately rather than jointly and assume pre-learned projections, which may not result in optimal performance. Among dictionary based methods, Yang $et$ $al.$ \cite{yang2012coupled} and Wang $et$ $al.$ \cite{wang2012semi} proposed learning dictionary pairs for cross modal synthesis. Qiu $et$ $al.$ \cite{qiu2012domain} proposed learning adaptive dictionaries for smooth domain shifts using regression. However, in practice, domain shifts are wide and often result in abrupt changes among features (eg., increase in resolution from a webcam image to a DSLR image). Shekhar $et$ $al.$ \cite{shekhar2013generalized} jointly projected the data onto a low dimensional space by preserving the manifold structure of the data from each domain, and learned a common adaptive dictionary for multiple domains, which can also be modified to learn discriminative dictionaries. However, the projected data may still possess a significant domain shift among the data distributions which may not result in an optimal solution. 


Considering the above challenges, we present a robust method that learns a common dictionary adapted to both source and target data. As the dimension of features may vary across the domains, we project the data onto a common low dimensional space by learning a projection matrix for each domain. In the process, we preserve the intrinsic geometry of the data from each domain and minimize the shift across the domains. Simultaneously, we learn an efficient and compact dictionary common to both the domains.  We extend our framework towards learning class specific discriminative dictionaries, as our final goal is classification. We additionally propose a discriminative manifold regularization, which imposes the intrinsic structure of class specific features onto the sparse coefficients to be obtained in the dictionary learning step. 

Our joint learning framework offers several advantages in terms of generalizability. First, learning domain specific projection matrices makes it easy to handle changes in feature dimensions. It also makes our algorithm  kernelizable. Second, learning the dictionary in a low dimensional space makes our algorithm faster and tractable. It also helps in discarding any redundant information present in the original features. Further, our method can be generalized to handle data from multiple domains. We present an efficient optimization approach to solve our problem, which has simple update steps.

The paper is organized in five sections. In Section 2, we formulate our dictionary learning framework, and the optimization scheme is described in Section 3. The evaluation approach using test data is described in Section 4. Experimental results are presented in Section 5. Section 6 concludes our work.


\section{Learning Framework}
The classic dictionary learning problem minimizes the representation error of the given data samples subject to sparsity constraint. Let $\mathbf{X} = [\mathbf{x}_1,\mathbf{x}_2,\cdots\mathbf{x}_N] \in \mathbb{R}^{n \times N}$ be the data matrix. Then the dictionary $\mathbf{D} \in \mathbb{R}^{n \times K}$ with $K$ atoms can be obtained by solving the following problem
\begin{align*}
\{\mathbf{D}^{*},\mathbf{S}^{*}\} = \argmin_{\mathbf{D},\mathbf{S}} \|\mathbf{X}-\mathbf{D}\mathbf{S}\|^{2}_{F} \hspace{0.05in} \mbox{s.t.} \hspace{0.05in} \|\mathbf{s}_i\|_{0} \hspace{0.05in} \leq T_{0} \hspace{0.05in} \forall i,
\end{align*}
where $\mathbf{S} = [\mathbf{s}_1,\mathbf{s}_2,\cdots\mathbf{s}_N] \in  \mathbb{R}^{K \times N}$ is a sparse representation matrix of $\mathbf{X}$ over $\mathbf{D}$ and $T_{0}$ is the sparsity level. The $\|\cdot\|_{0}$-norm counts the number of nonzero elements in a vector and $\|\cdot\|_{F}$ is the Frobenius norm of a matrix.

We consider a case where we have data from two domains, $\mathbf{X}_1 \in \mathbb{R}^{n_1 \times N_1}$ and $\mathbf{X}_2 \in \mathbb{R}^{n_2 \times N_2}$. Our goal is to find projection matrices $\mathbf{P}_1 \in \mathbb{R}^{n_1 \times n}$ and $\mathbf{P}_2 \in \mathbb{R}^{n_1 \times n}$ which map $\mathbf{X}_1$ and $\mathbf{X}_2$ onto a low-dimensional space and simultaneously learn a common dictionary $\mathbf{D} \in \mathbb{R}^{n \times K}$ for both the domains. We enforce orthonormality constraint on columns of projection matrices $\mathbf{P}_1$ and $\mathbf{P}_2$, in order to prevent the solution from becoming degenerate. We will see later that, this particular assumption paves way for an efficient optimization approach.

While bringing the data from two domains to a low dimensional space, it is desirable that the projections preserve much of the information which is available in the original domains. To facilitate such preservation, we wish to minimize the following cost function which includes a manifold regularization \cite{belkin2006manifold} term for data from each domain:
\begin{align*}
\begin{split}
\mathcal{C}_1(\mathbf{P}_1,\mathbf{P}_2) = tr(\mathbf{P}_1^T\mathbf{X}_1 \mathbf{L}_1  \mathbf{X}_1^{T}\mathbf{P}_1) + tr(\mathbf{P}_2^T\mathbf{X}_2 \mathbf{L}_2  \mathbf{X}_2^{T}\mathbf{P}_2),
\end{split}
\end{align*}
where $tr(\cdot)$ is the trace of a matrix and $\mathbf{L}_1$, $\mathbf{L}_2$ are the normalized graph-Laplacian matrices associated with the nearest neighborhood graphs constructed from data matrices $\mathbf{X}_1$, $\mathbf{X}_2$ respectively.

The above cost function enforces the condition that, if two points each domain are close to each other in the original space, they are required to be closer to each other in the projected space as well. This assumption is known as manifold assumption \cite{belkin2006manifold}, which has been used successfully for non-linear dimensionality reduction and semi-supervised learning techniques \cite{belkin2006manifold}.

To make the learned dictionary adaptive to both the domains, it should capture the commonality among the domains. But the data among the domains will have largely different distributions. So, there will be a large domain shift among the data even in the reduced space. We seek to minimize this domain shift. To realize this, a natural strategy is to make the data distributions of both the domains as close as possible. In our work, we follow \cite{gretton2006kernel,pan2011domain,long2013transfer} and use the Maximum Mean Discrepancy (MMD) as the distance measure between the data distributions. It computes the distance between the sample means of both the distributions:
\begin{align*}
\mathcal{C}_2(\mathbf{P}_1,\mathbf{P}_2) = \left\|\frac{1}{N_1}\sum_{i=1}^{N_1} \mathbf{P}_1^T  \mathbf{x}_{1i}- \frac{1}{N_2}\sum_{j=1}^{N_2}\mathbf{P}_2^T \mathbf{x}_{2j}\right\|^2.
\end{align*}
After projecting the data onto the common low dimensional space, we seek to minimize the following representation error:
\begin{align*}
\begin{split}
\mathcal{C}_3(\mathbf{D},\mathbf{P}_1,\mathbf{P}_2,\mathbf{S}_1,\mathbf{S}_2) = \|\mathbf{P}_1^T\mathbf{X}_1-\mathbf{D}\mathbf{S}_1\|^{2}_{F} + \\ \|\mathbf{P}_2^T\mathbf{X}_2-\mathbf{D}\mathbf{S}_2\|^{2}_{F}
\end{split}
\end{align*}
The above costs $\mathcal{C}_1$, $\mathcal{C}_2$, $\mathcal{C}_3$ can be rewritten in block-matrix form as:
\begin{align}\label{eq1}
\begin{split}
& \mathcal{C}_1(\tilde{\mathbf{P}}) = tr(\tilde{\mathbf{P}}^T\tilde{\mathbf{X}} \tilde{\mathbf{L}}  \tilde{\mathbf{X}}^{T}\tilde{\mathbf{P}})\\
& \mathcal{C}_2(\tilde{\mathbf{P}}) = tr(\tilde{\mathbf{P}}^T\tilde{\mathbf{X}} \tilde{\mathbf{M}}  \tilde{\mathbf{X}}^{T}\tilde{\mathbf{P}})\\
& \mathcal{C}_3(\mathbf{D},\tilde{\mathbf{P}},\tilde{\mathbf{S}}) = 
\|\tilde{\mathbf{P}}^T\tilde{\mathbf{X}}-\mathbf{D}\tilde{\mathbf{S}}\|^{2}_{F}
\end{split}
\end{align}
where
$\tilde{\mathbf{P}}^T = [\mathbf{P}_1^T \hspace{0.1in} \mathbf{P}_2^T]$, $\tilde{\mathbf{X}} = diag(\mathbf{X}_1,\mathbf{X}_2)$ and $\tilde{\mathbf{S}} = [\mathbf{S}_1 \hspace{0.1in} \mathbf{S}_2]$. Here, $diag$ denotes the block diagonal matrix formed from the data matrices $\mathbf{X}_1$ and $\mathbf{X}_2$. The MMD matrix $\mathbf{M}$ is computed as:
\begin{align}\label{eq2}
\mathbf{M}_{ij} = \begin{cases} 1/N_1^2, &\tilde{\mathbf{x}}_i,\tilde{\mathbf{x}}_j \in \mathbf{X}_1 \\
1/N_2^2, &\tilde{\mathbf{x}}_i,\tilde{\mathbf{x}}_j \in \mathbf{X}_2 \\
-\frac{1}{N_1 N_2}, &\tilde{\mathbf{x}}_i \in \mathbf{X}_1, \tilde{\mathbf{x}}_j \in \mathbf{X}_2 \end{cases}
\end{align}
 The overall optimization is given as:
\begin{align}\label{eq3}
\begin{split}
 \{\mathbf{D}^{*},\tilde{\mathbf{P}}^{*},\tilde{\mathbf{S}}^{*}\} = \argmin_{\mathbf{D},\tilde{\mathbf{P}},\tilde{\mathbf{S}}} \mathcal{C}_3(\mathbf{D},\tilde{\mathbf{P}},\tilde{\mathbf{S}}) \\ + \lambda_1 \mathcal{C}_1(\tilde{\mathbf{P}}) + \lambda_2 \mathcal{C}_2(\tilde{\mathbf{P}}) \\
 \mbox{s.t.} \hspace{0.05in} \mathbf{P}_i^T\mathbf{P}_i = I, \hspace{0.05in} i = 1,2 \hspace{0.05in} \mbox{and} \hspace{0.05in} \|\tilde{\mathbf{s}}_j\|_{0} \hspace{0.05in} \leq T_{0} \hspace{0.05in} \forall j
\end{split}
\end{align}

The above formulation can be conveniently extended to multiple domains. For an $m$ domain problem, the block matrices can be constructed as $\tilde{\mathbf{P}}^T = [\mathbf{P}_1^T \hspace{0.1in} \mathbf{P}_2^T \cdots \mathbf{P}_m^T]$, $\tilde{\mathbf{X}} = diag(\mathbf{X}_1, \mathbf{X}_2 \cdots \mathbf{X}_m)$ and $\tilde{\mathbf{S}} = [\mathbf{S}_1 \hspace{0.1in} \mathbf{S}_2 \cdots \mathbf{S}_m]$.

\subsection{Manifold Regularization}
To make the atoms of the dictionary respect the intrinsic structures of data, Cai $et$ $al.$ \cite{zheng2011graph} proposed a Graph Regularized Sparse Coding (GraphSC) method, which further explores the manifold assumption \cite{belkin2006manifold}. GraphSC assumes that if two points $\tilde{\mathbf{x}}_i$ and $\tilde{\mathbf{x}}_j$ are close in the intrinsic geometry of data on the projected space, then their sparse representations $\tilde{\mathbf{s}}_i$ and $\tilde{\mathbf{s}}_j$ are also close. Adding this regularization to the cost $\mathcal{C}_3$:
\begin{align}\label{eq4}
\mathcal{C}_3(\mathbf{D},\tilde{\mathbf{P}},\tilde{\mathbf{S}}) = 
\|\tilde{\mathbf{P}}^T\tilde{\mathbf{X}}-\mathbf{D}\tilde{\mathbf{S}}\|^{2}_{F} + \lambda_3 tr(\tilde{\mathbf{S}}\mathbf{L}_p\tilde{\mathbf{S}}^T),
\end{align}
where $\mathbf{L}_p$ is the normalized graph-Laplacian associated with the nearest neighborhood graph formed from the data $\tilde{\mathbf{P}}^T\tilde{\mathbf{X}}$ in the projected space.

\subsection{Discriminative Dictionaries}
The dictionary learned using above approach can reconstruct data from multiple domains well, but it cannot discriminate among the data from different classes. Following recent advances \cite{ramirez2010classification,yang2011fisher} in learning discriminative dictionaries, we split the dictionary $\mathbf{D}$ into class specific dictionaries $\{\mathbf{D}_1,\cdots \mathbf{D}_C\}$, where $C$ is the total number of classes. We modify the cost function $\mathcal{C}_3$ as:
\begin{align}\label{eq5}
\begin{split}
\mathcal{C}_3(\mathbf{D},\tilde{\mathbf{P}},\tilde{\mathbf{S}}) = 
\|\tilde{\mathbf{P}}^T\tilde{\mathbf{X}}-\mathbf{D}\tilde{\mathbf{S}}\|^{2}_{F} + \mu_1 \|\tilde{\mathbf{P}}^T\tilde{\mathbf{X}}-\mathbf{D}\tilde{\mathbf{S}}_{in}\|^{2}_{F} \\ + \mu_2 \|\mathbf{D}\tilde{\mathbf{S}}_{out}\|^{2}_{F} + \lambda_3 tr(\tilde{\mathbf{S}}_{in}\mathbf{L}_p\tilde{\mathbf{S}}_{in}^T),
\end{split}
\end{align}
where the weights $\mu_1$ and $\mu_2$ influence the discriminative power of the dictionary $\mathbf{D}$. The matrices $\tilde{\mathbf{S}}_{in}$ and $\tilde{\mathbf{S}}_{out}$ are given as:
\begin{align*}
\tilde{\mathbf{S}}_{in}(i,j) = \begin{cases} \tilde{\mathbf{S}}(i,j) & \mathbf{d}_i, \tilde{\mathbf{y}}_j \in \mbox{same class} \\
0 & \mbox{otherwise} \end{cases}
\end{align*}
and
\begin{align*}
\tilde{\mathbf{S}}_{out}(i,j) = \begin{cases} \tilde{\mathbf{S}}(i,j) & \mathbf{d}_i, \tilde{\mathbf{y}}_j \in \mbox{different class} \\
0 & \mbox{otherwise} \end{cases}
\end{align*}
This way, we learn the dictionary of a particular class one at a time, i.e. \eqref{eq5} encourages the reconstruction of a dictionary of the corresponding class and penalizes the reconstruction of the dictionaries of other classes. We note that the manifold regularization term in \eqref{eq5} is now aware of discrimination, as it handles only the data from the corresponding class and omits the data from other classes. 

The cost function $\mathcal{C}_3$ in \eqref{eq5} can handle only labeled data from each domain. Unlabeled data can be handled using semi-supervised approaches such as \cite{shrivastava2012learning}, which is beyond the scope of this paper.

\subsection{Kernelization}
Due to the non-linear structure of the data, projecting the original features may not be efficient. To overcome this drawback, we map the original features onto a high dimensional space before projecting them. Let $\Phi: \mathbb{R}^{n_i} \mapsto \mathcal{H}$ be a mapping from the space of domain $i$ to the reproducing kernel Hilbert space $\mathcal{H}$. The projection $\mathbf{P}_i:\mathbb{R}^n \mapsto \mathcal{H}$ which maps to the low dimensional space be a compact linear operator. Let $\tilde{\mathbf{K}} = \langle \Phi(\tilde{\mathbf{X}}),\Phi(\tilde{\mathbf{X}})\rangle_{\mathcal{H}}$ be the kernel matrix associated with $\mathcal{H}$. The representer theorem \cite{scholkopf2002learning} states that $\mathbf{P}_i$ can be represented as 
\begin{align*}
\mathbf{P}_i = \Phi(\mathbf{X}_i) \mathbf{A}_i 
\end{align*}
for some matrix $\mathbf{A}_i \in \mathbb{R}^{N_i \times n}$. Using the above expression for projection matrices, we redefine the cost functions and the equality constraints as 

\begin{align} \label{eq6}
\begin{split}
& \mathcal{C}_1(\tilde{\mathbf{A}}) = tr(\tilde{\mathbf{A}}^T\tilde{\mathbf{K}} \tilde{\mathbf{L}}  \tilde{\mathbf{K}}^{T}\tilde{\mathbf{A}})\\
& \mathcal{C}_2(\tilde{\mathbf{A}}) = tr(\tilde{\mathbf{A}}^T \tilde{\mathbf{K}} \tilde{\mathbf{M}}  \tilde{\mathbf{K}}^{T}\tilde{\mathbf{A}})\\
& \mathcal{C}_3(\mathbf{D},\tilde{\mathbf{A}},\tilde{\mathbf{S}}) = 
\|\tilde{\mathbf{A}}^T\tilde{\mathbf{K}}-\mathbf{D}\tilde{\mathbf{S}}\|^{2}_{F} + \mu_1 \|\tilde{\mathbf{A}}^T\tilde{\mathbf{K}}-\mathbf{D}\tilde{\mathbf{S}}_{in}\|^{2}_{F} \\ & + \mu_2 \|\mathbf{D}\tilde{\mathbf{S}}_{out}\|^{2}_{F} + \lambda_3 tr(\tilde{\mathbf{S}}_{in}\mathbf{L}_p\tilde{\mathbf{S}}_{in}^T)\\
& \mbox{s.t.} \hspace{0.05in} \mathbf{A}_i^T\mathbf{K}_i\mathbf{A}_i = I, \hspace{0.1in} \mathbf{K}_i = \langle \Phi(\mathbf{X}_i),\Phi(\mathbf{X}_i)\rangle_{\mathcal{H}} \\ & \forall i = 1,\cdots M.
\end{split}
\end{align}

\section{Optimization}
The above optimization problem \eqref{eq6} is non-convex in $\mathbf{D}$, $\tilde{\mathbf{A}}$ and $\tilde{\mathbf{S}}$. We solve it in iterative alternating steps. At each iteration, three update steps are performed namely projection update, dictionary update and sparse code update.

\subsection{Projection Update}
In this step, we update $\tilde{\mathbf{A}}$ by assuming $\mathbf{D}$ and $\tilde{\mathbf{S}}$ are fixed. Due to the orthonormality constraint on projection matrices, this step involves optimization on the Stiefel manifold. We solved this problem using the efficient approach presented in \cite{shekhar2013generalized,wen2013feasible}.

\subsection{Dictionary and Sparse code Update}
When $\tilde{\mathbf{A}}$ is fixed, this problem boils down to a discriminative dictionary learning with the data matrix as $\mathbf{Z} = \tilde{\mathbf{A}}^T\tilde{\mathbf{K}}$. We use the discriminative dictionary learning approach presented in \cite{yang2011fisher} to update $\mathbf{D}$ and $\tilde{\mathbf{S}}$.

\section{Test Evaluation}
As our goal is classification, given a test sample $\mathbf{x}_t$ from the domain $i$, we propose the following steps, similar to \cite{shekhar2013generalized,nguyen2012sparse}. We map the sample into kernel space $\Phi(\mathbf{x}_t)$.
\begin{enumerate}
\item Compute the low dimensional embedding $\mathbf{z}_t$ of the sample, using the projection matrix $\mathbf{P}^*_i$,
\begin{align*}
\mathbf{z}_t = \mathbf{P}^{*T}_i\Phi(\mathbf{x}_t) = \mathbf{A}_i^T \mathbf{K}_t
\end{align*}
where $\mathbf{K}_t = \langle \Phi(\mathbf{X}_i),\Phi(\mathbf{x}_t)\rangle_{\mathcal{H}}$
\item Compute the sparse code $\bar{\mathbf{s}}_t$ of the embedded test sample over the dictionary $\mathbf{D}$ using the OMP algorithm \cite{pati1993orthogonal}
\begin{align*}
\bar{\mathbf{s}}_t = \argmin_{\mathbf{s}} \|\mathbf{x}_t - \mathbf{D}\mathbf{s}\|^2_F \hspace{0.1in} \mbox{s.t.} \hspace{0.1in} \|\mathbf{s}\|_0 \leq T_0
\end{align*}
\item The test sample can now be allocated to class $c$, if the reconstruction error using the class specific dictionary $\mathbf{D}_c$ and the corresponding sparse code $\bar{\mathbf{s}}_t^c$ is minimum. For a better discriminative results, it is desired to compute the error in the original feature space rather than the low dimensional space. So, we map the dictionary $\mathbf{D}_c$ onto $\mathcal{H}$ and allocate the test sample as:
\begin{align*}
\mbox{Output class} = \argmin_{c = 1, \cdots C} \|\Phi(\mathbf{x}_t) - \mathbf{P}^*_i \mathbf{D}_c \bar{\mathbf{s}}_t^c\|^2_F
\end{align*}
\end{enumerate}

\section{Experiments}
We conduct experiments on image classification to validate the effectiveness of our proposed method. We show the performance of our method on two adaptation databases and compare it with the existing state-of-the-art adaptation algorithms. For each database, the results are averaged over 20 runs of random train/test splits.

\subsection{Office and Caltech datasets}

\begin{figure}
\begin{center}
\begin{tabular}{cccc}
\includegraphics[width=45pt]{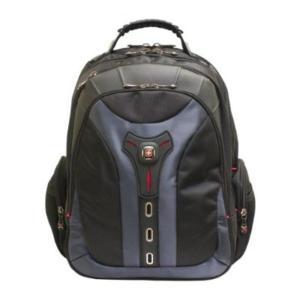} &
\includegraphics[width=45pt]{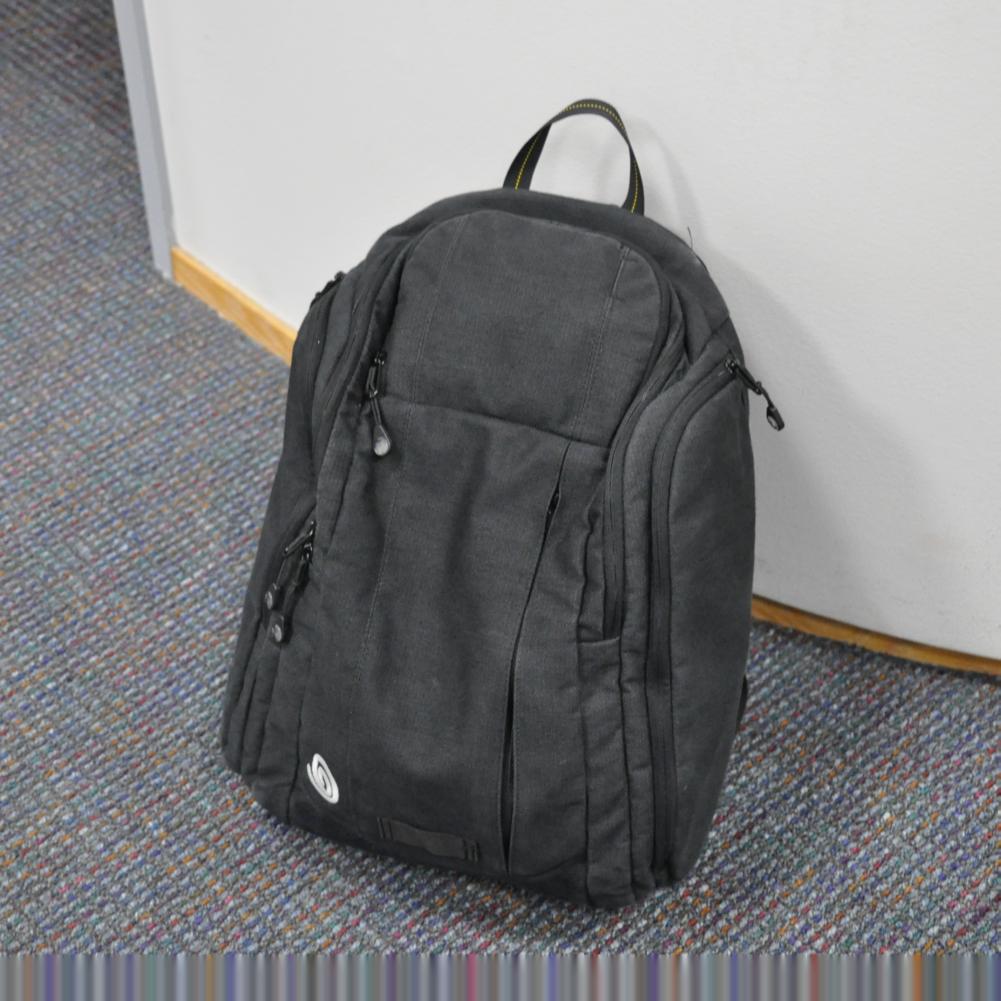} &
\includegraphics[width=45pt]{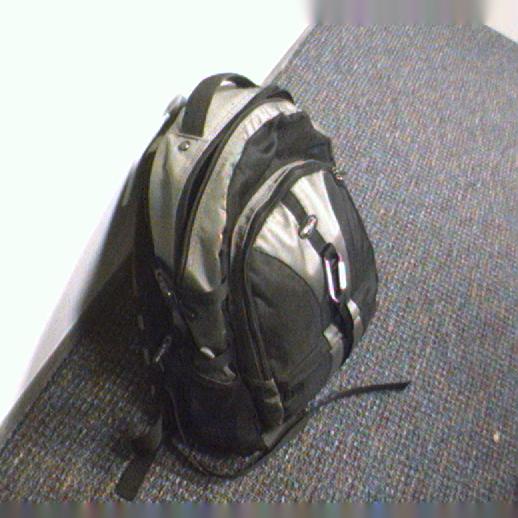} &
\includegraphics[width=45pt]{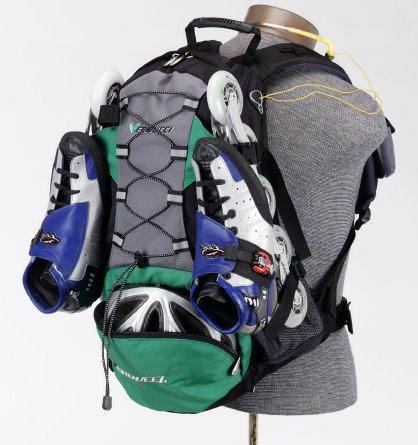} \\
\includegraphics[width=45pt]{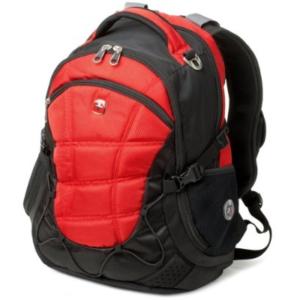} &
\includegraphics[width=45pt]{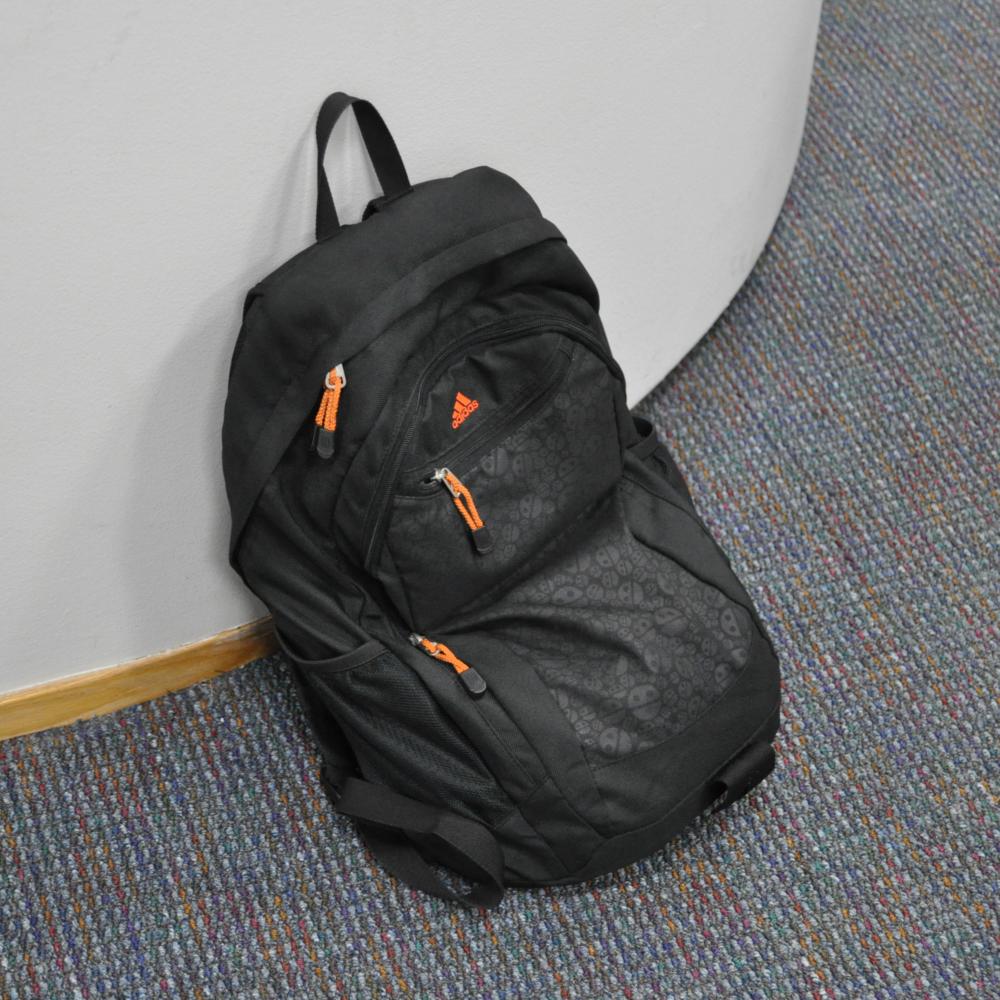} &
\includegraphics[width=45pt]{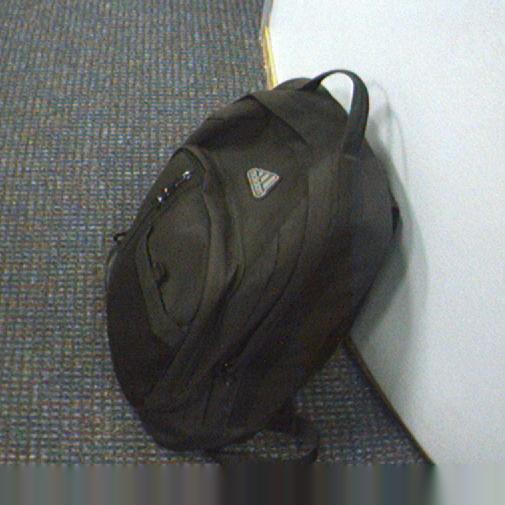} &

\includegraphics[width=45pt]{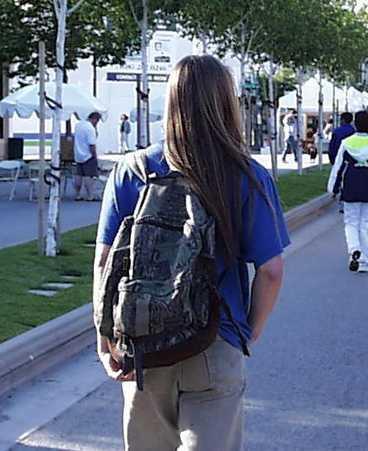} \\
(a)&(b)&(c)&(d)
\end{tabular}
\end{center}
\caption{ some backpack images of (a) Amazon, (b) DSLR, (c) Webcam \& (d) Caltech-256.
\label{fig:1}}
\end{figure}

Office \cite{saenko2010adapting} is a popular benchmark dataset used for visual domain adaptation. The dataset contains three domains of images namely, Amazon which consists of the images downloaded from online merchants, DSLR consists of high resolution images, Webcam consists of low resolution images. It has $4,652$ images and $31$ classes. In addition, we choose the Caltech-256 dataset \cite{griffin2007caltech} as the fourth domain. Fig. (\ref{fig:1}) shows some BACKPACK images of all the four domains. We choose two different scenarios to test our algorithm. In the first scenario, we use 10 classes common to all four domains: BACKPACK, TOURING-BIKE, CALCULATOR, HEADPHONES, COMPUTER-KEYBOARD, LAPTOP, COMPUTER-MONITOR, COMPUTER-MOUSE, COFFEE-MUG and VIDEO-PROJECTOR. There are a total of 2533 images in this scenario with 8 to 151 images in each class. In the second scenario, we restrict to the office dataset and test on all the 31 classes in it. In this scenario, we test our method using multiple domains. In both the scenarios, we use 20 samples per class for Amazon/Caltech and 8 samples per class for Webcam/DSLR when used as a source domain. We use 3 samples per class for all the four domains when used as the target for testing. We compare our results with those obtained from \cite{saenko2010adapting,gong2012geodesic,yang2011fisher,gopalan2011domain,
shekhar2013generalized}.

\begin{table*}\label{table1}
\centering
\caption{Performance comparison on four domains (C: caltech, A: amazon, D: dslr, W: webcam) used as a single source}
\begin{tabular}{|c|c|c|c|c|c|c|c|c|}
\hline
Methods & C $\rightarrow$ A & C $\rightarrow$ D & A $\rightarrow$ C & A $\rightarrow$ W & W $\rightarrow$ C & W $\rightarrow$ A & D $\rightarrow$ A & D $\rightarrow$ W\\
\hline
Metric \cite{saenko2010adapting} & 33.7 $\pm$ 0.8  & 35.0 $\pm$ 1.1 & 27.3 $\pm$ 0.7 & 36.0 $\pm$ 1.0 & 21.7 $\pm$ 0.5 & 32.3 $\pm$ 0.8 & 32.0 $\pm$ 0.8 & 55.6 $\pm$ 0.7\\
SGF \cite{gopalan2011domain} & 40.2 $\pm$ 0.7  & 36.6 $\pm$ 0.8 & 37.7 $\pm$ 0.5 & 37.9 $\pm$ 0.7 & 29.2 $\pm$ 0.7 & 38.2 $\pm$ 0.6 & 39.2 $\pm$ 0.7 & 69.5 $\pm$ 0.9\\
GFK \cite{gong2012geodesic} & 46.1 $\pm$ 0.6  & 55.0 $\pm$ 0.9 & \bf{39.6 $\pm$} 0.4 & 56.9 $\pm$ 1.0 & 32.8 $\pm$ 0.1 & 46.2 $\pm$ 0.6 & 46.2 $\pm$ 0.6 & \bf{80.2 $\pm$ 0.4}\\
FDDL \cite{yang2011fisher} & 39.3 $\pm$ 2.9  & 55.0 $\pm$ 2.8 & 24.3 $\pm$ 2.2 & 50.4 $\pm$ 3.5 & 22.9 $\pm$ 2.6 & 41.1 $\pm$ 2.6 & 36.7 $\pm$ 2.5 & 65.9 $\pm$ 4.9\\
SDDL \cite{shekhar2013generalized} & 49.5 $\pm$ 2.6  & 76.7 $\pm$ 3.9 & 27.4 $\pm$ 2.4 & 72.0 $\pm$ 4.8 & 29.7 $\pm$ 1.9 & 49.4 $\pm$ 2.1 & 48.9 $\pm$ 3.8 & 72.6 $\pm$ 2.1\\
Ours & \bf{52.8 $\pm$ 3.6}  & \bf{79.7 $\pm$ 4.9} & 29.1 $\pm$ 2.6 & \bf{74.9 $\pm$ 5.0} & \bf{33.1 $\pm$ 2.7} & \bf{53.1 $\pm$ 4.0} & \bf{52.2 $\pm$ 4.4} & 77.5 $\pm$ 3.5\\
\hline
\end{tabular}
\end{table*}

\begin{table*}\label{table2}
\centering
\caption{Performance comparison on multiple domains among amazon, webcam and dslr used for source data}
\begin{tabular}{|c|c|c|c|c|c|c|}
\hline
Source & Target & SGF \cite{gopalan2011domain} & RDALR \cite{jhuo2012robust} & FDDL \cite{yang2011fisher} & SDDL\cite{shekhar2013generalized} & Ours \\
\hline
dslr, amazon & webcam & 52 $\pm$ 2.5  & 36.9 $\pm$ 1.1 & 41.0 $\pm$ 2.4 & 57.8 $\pm$ 2.4 & \bf{60.2 $\pm$ 3.5}\\
amazon, webcam & dslr & 39 $\pm$ 1.1  & 31.2 $\pm$ 1.3 & 38.4 $\pm$ 3.4 & 56.7 $\pm$ 2.3 & \bf{58.4 $\pm$ 3.2}\\
webcam, dslr & amazon & \bf{28 $\pm$ 0.8}  & 20.9 $\pm$ 0.9 & 19.0 $\pm$ 1.2 & 24.1 $\pm$ 1.6 & 26.2 $\pm$ 2.2\\
\hline
\end{tabular}
\end{table*}

\paragraph{Features for images.} We used the 800 bin SURF features provided by \cite{saenko2010adapting} for Amazon, Webcam and DSLR domains. For the Caltech domain, the 800 bin SURF features provided by \cite{shekhar2013generalized} are used.

\paragraph{Parameter settings.} We used the non-parametric histogram intersection kernel in all our experiments. We set $\mu_1 = 4$, $\mu_2 = 30$ $\lambda_1 = 1$, $\lambda_2 = 50$ and $\lambda_3 = 1$ for our experiments. For the first scenario, we choose to learn 4 dictionary atoms per class, i.e. $K = 40$ for ten classes and the final dimension $n = 60$. For the second scenario, we choose 6 dictionary atoms per class, i.e. $K = 186$ for thirty one classes and $n = 90$. For SDDL \cite{shekhar2013generalized} and FDDL \cite{yang2011fisher}, we fix the parameters as given in \cite{shekhar2013generalized} as they are found to give the best results.

\subsubsection{Results using single source}
The comparison of our results with those obtained from other methods is shown in Table 1. Our algorithm performs best for 6 domain pairs and second best for 1 pair. Further, we can see that our method outperforms SDDL among all the domain pairs. So, we can infer that our domain shift minimizing framework improves the efficiency over \cite{shekhar2013generalized}, specifically when the training data and test data come from different distributions.

\subsubsection{Results using multiple sources}
We performed experiments using multiple domains by choosing among Amazon, Webcam and DSLR as sources. Table 2 shows the three possible combinations and their results. Our results outperform those of SGF \cite{gopalan2011domain} in two cases and SDDL \cite{shekhar2013generalized} in all the cases.

\subsection{USPS and MNIST datasets}
\begin{figure}
\begin{center}
\begin{tabular}{c}
\includegraphics[width=180pt]{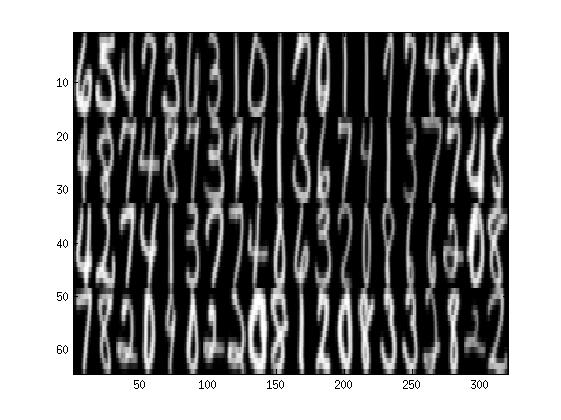} \\
\includegraphics[width=180pt]{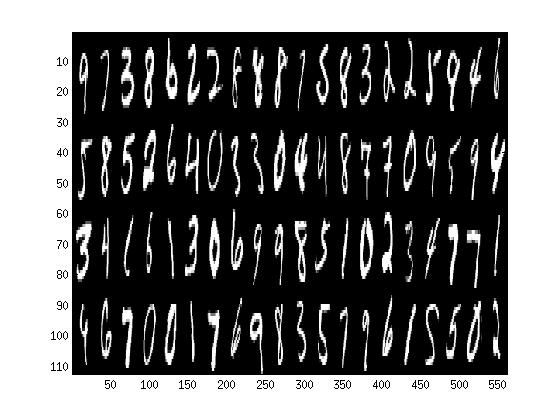} \\
\end{tabular}
\end{center}
\caption{some of the USPS (first row) and MNIST (second row) handwritten digit images
\label{fig:2}}
\end{figure}

The USPS and MNIST are handwritten digit image datasets used widely for digit recognition, classification etc. The USPS dataset consists of $7,291$ training images and $2007$ test images of size $16 \times 16$. MNIST dataset has a training set of $60,000$ images and a test set of $10,000$ images each of size $28 \times 28$. Some of the images of both the datasets are shown in Fig (\ref{fig:2}). For our experiments, we adopt the publicly available USPS+MNIST datasets provided by Long $et$ $al.$ \cite{long2013transfer}. The datasets contain $1800$ USPS and $2000$ MNIST images of 10 classes. All the images are scaled to $16 \times 16$, and each is represented by a $256 \times 1$ vector which encodes the gray level values. For each domain of this database, we use 20 samples per class when used as a source and 3 samples per class when used as a target.  We use the same kernel and the set of parameters which are used for the earlier database. We choose to learn 4 dictionary atoms per class, i.e. $K = 40$ for ten classes and the final dimension $n = 60$ for this database. We compare the performance of our method with those obtained from \cite{yang2011fisher,shekhar2013generalized}.

\subsubsection{Results}
Table 3 shows the comparison of our results with those of other methods. We evaluated our method considering USPS as source, MNIST as the target and vice-versa. We can see that results using our approach outperform those obtained from the other methods.

\begin{table}\label{table3}
\centering
\caption{Performance comparison on U: USPS and M: MNIST as source domains}
\begin{tabular}{|c|c|c|c|c|}
\hline
Source & Target & FDDL \cite{yang2011fisher} & SDDL\cite{shekhar2013generalized} & Ours \\
\hline
U & M & 58.9 $\pm$ 2.1  & 61.1 $\pm$ 2.4 & \bf{65.6 $\pm$ 3.4} \\
M & U & 69.2 $\pm$ 3.6  & 72.2 $\pm$ 3.3 & \bf{75.3 $\pm$ 3.7}  \\
\hline
\end{tabular}
\end{table}

\section{Conclusion}
We presented a generalized framework for adapting dictionaries to multiple domains by minimizing the domain shift. Furthermore, we showed that the method can be kernelized and can be modified to learn discriminative dictionaries for class specific data. The dictionary is learned on a common low dimensional space, on which the original data is projected. We show that our method outperforms the current state-of-the-art methods on different adaptation databases. Future works include finding a way to leverage the unlabeled data while training and to implement tractable, online adaptations of dictionaries, for large-scale data. 
{\small
\bibliographystyle{ieee}
\bibliography{WACV2}

\begin{thebibliography}{10}\itemsep=-1pt

\bibitem{belkin2006manifold}
M.~Belkin, P.~Niyogi, and V.~Sindhwani.
\newblock Manifold regularization: A geometric framework for learning from
  labeled and unlabeled examples.
\newblock {\em The Journal of Machine Learning Research}, 7:2399--2434, 2006.

\bibitem{daume2009frustratingly}
H.~Daum{\'e}~III.
\newblock Frustratingly easy domain adaptation.
\newblock {\em arXiv preprint arXiv:0907.1815}, 2009.

\bibitem{elad2006image}
M.~Elad and M.~Aharon.
\newblock Image denoising via sparse and redundant representations over learned
  dictionaries.
\newblock {\em Image Proc., IEEE Trans. on}, 15(12):3736--3745, 2006.

\bibitem{gong2012geodesic}
B.~Gong, Y.~Shi, F.~Sha, and K.~Grauman.
\newblock Geodesic flow kernel for unsupervised domain adaptation.
\newblock In {\em CVPR}, pages 2066--2073. IEEE, 2012.

\bibitem{gopalan2011domain}
R.~Gopalan, R.~Li, and R.~Chellappa.
\newblock Domain adaptation for object recognition: An unsupervised approach.
\newblock In {\em ICCV}, pages 999--1006. IEEE, 2011.

\bibitem{gretton2006kernel}
A.~Gretton, K.~M. Borgwardt, M.~Rasch, B.~Sch{\"o}lkopf, and A.~J. Smola.
\newblock A kernel method for the two-sample-problem.
\newblock In {\em NIPS}, pages 513--520, 2006.

\bibitem{griffin2007caltech}
G.~Griffin, A.~Holub, and P.~Perona.
\newblock Caltech-256 object category dataset.
\newblock 2007.

\bibitem{han2012sparse}
Y.~Han, F.~Wu, D.~Tao, J.~Shao, Y.~Zhuang, and J.~Jiang.
\newblock Sparse unsupervised dimensionality reduction for multiple view data.
\newblock {\em Circuits and Sys. for Video Tech., IEEE Trans. on},
  22(10):1485--1496, 2012.

\bibitem{jhuo2012robust}
I.-H. Jhuo, D.~Liu, D.~Lee, and S.-F. Chang.
\newblock Robust visual domain adaptation with low-rank reconstruction.
\newblock In {\em CVPR}, pages 2168--2175. IEEE, 2012.

\bibitem{jia2010factorized}
Y.~Jia, M.~Salzmann, and T.~Darrell.
\newblock Factorized latent spaces with structured sparsity.
\newblock In {\em NIPS}, pages 982--990, 2010.

\bibitem{kulis2011you}
B.~Kulis, K.~Saenko, and T.~Darrell.
\newblock What you saw is not what you get: Domain adaptation using asymmetric
  kernel transforms.
\newblock In {\em CVPR}, pages 1785--1792. IEEE, 2011.

\bibitem{long2013transfer}
M.~Long, J.~Wang, G.~Ding, J.~Sun, and P.~Yu.
\newblock Transfer joint matching for unsupervised domain adaptation.
\newblock In {\em Proc. of IEEE CVPR}, pages 1410--1417, 2013.

\bibitem{nguyen2012sparse}
H.~V. Nguyen, V.~M. Patel, N.~M. Nasrabadi, and R.~Chellappa.
\newblock Sparse embedding: A framework for sparsity promoting dimensionality
  reduction.
\newblock In {\em ECCV}, pages 414--427. Springer, 2012.

\bibitem{olshausen1997sparse}
B.~A. Olshausen and D.~J. Field.
\newblock Sparse coding with an overcomplete basis set: A strategy employed by
  v1?
\newblock {\em Vision research}, 37(23):3311--3325, 1997.

\bibitem{pan2011domain}
S.~J. Pan, I.~W. Tsang, J.~T. Kwok, and Q.~Yang.
\newblock Domain adaptation via transfer component analysis.
\newblock {\em Neural Networks, IEEE Trans. on}, 22(2):199--210, 2011.

\bibitem{pati1993orthogonal}
Y.~C. Pati, R.~Rezaiifar, and P.~Krishnaprasad.
\newblock Orthogonal matching pursuit: Recursive function approximation with
  applications to wavelet decomposition.
\newblock In {\em Signals, Systems and Computers, 1993. 1993 Conference Record
  of The Twenty-Seventh Asilomar Conference on}, pages 40--44. IEEE, 1993.

\bibitem{qiu2012domain}
Q.~Qiu, V.~M. Patel, P.~Turaga, and R.~Chellappa.
\newblock Domain adaptive dictionary learning.
\newblock In {\em ECCV}, pages 631--645. Springer, 2012.

\bibitem{ramirez2010classification}
I.~Ramirez, P.~Sprechmann, and G.~Sapiro.
\newblock Classification and clustering via dictionary learning with structured
  incoherence and shared features.
\newblock In {\em CVPR}, pages 3501--3508. IEEE, 2010.

\bibitem{saenko2010adapting}
K.~Saenko, B.~Kulis, M.~Fritz, and T.~Darrell.
\newblock Adapting visual category models to new domains.
\newblock In {\em ECCV}, pages 213--226. Springer, 2010.

\bibitem{scholkopf2002learning}
B.~Sch{\"o}lkopf and A.~J. Smola.
\newblock {\em Learning with kernels}.
\newblock “The” MIT Press, 2002.

\bibitem{shekhar2013generalized}
S.~Shekhar, V.~M. Patel, H.~V. Nguyen, and R.~Chellappa.
\newblock Generalized domain-adaptive dictionaries.
\newblock In {\em CVPR}, pages 361--368. IEEE, 2013.

\bibitem{shrivastava2012learning}
A.~Shrivastava, J.~K. Pillai, V.~M. Patel, and R.~Chellappa.
\newblock Learning discriminative dictionaries with partially labeled data.
\newblock In {\em ICIP}, pages 3113--3116. IEEE, 2012.

\bibitem{wang2012semi}
S.~Wang, D.~Zhang, Y.~Liang, and Q.~Pan.
\newblock Semi-coupled dictionary learning with applications to image
  super-resolution and photo-sketch synthesis.
\newblock In {\em CVPR}, pages 2216--2223. IEEE, 2012.

\bibitem{wen2013feasible}
Z.~Wen and W.~Yin.
\newblock A feasible method for optimization with orthogonality constraints.
\newblock {\em Mathematical Programming}, 142(1-2):397--434, 2013.

\bibitem{wright2009robust}
J.~Wright, A.~Y. Yang, A.~Ganesh, S.~S. Sastry, and Y.~Ma.
\newblock Robust face recognition via sparse representation.
\newblock {\em Pat. Analy. and Mach. Int., IEEE Trans. on}, 31(2):210--227,
  2009.

\bibitem{yang2012coupled}
J.~Yang, Z.~Wang, Z.~Lin, S.~Cohen, and T.~Huang.
\newblock Coupled dictionary training for image super-resolution.
\newblock {\em Image Proc., IEEE Trans. on}, 21(8):3467--3478, 2012.

\bibitem{yang2011fisher}
M.~Yang, D.~Zhang, and X.~Feng.
\newblock Fisher discrimination dictionary learning for sparse representation.
\newblock In {\em ICCV}, pages 543--550. IEEE, 2011.

\bibitem{zheng2011graph}
M.~Zheng, J.~Bu, C.~Chen, C.~Wang, L.~Zhang, G.~Qiu, and D.~Cai.
\newblock Graph regularized sparse coding for image representation.
\newblock {\em Image Proc., IEEE Trans. on}, 20(5):1327--1336, 2011.

\end{thebibliography}
}

\end{document}